\documentclass[conference]{IEEEtran}
\IEEEoverridecommandlockouts
\usepackage{cite}
\usepackage{amsmath,amssymb,amsfonts}
\usepackage{algorithmic}
\usepackage{graphicx}
\usepackage{textcomp}
\usepackage{tabularx}
\usepackage{ltxtable}
\usepackage[table]{xcolor}  
\usepackage{balance}
\usepackage{caption}
\usepackage{subcaption}

\def\BibTeX{{\rm B\kern-.05em{\sc i\kern-.025em b}\kern-.08em
    T\kern-.1667em\lower.7ex\hbox{E}\kern-.125emX}}

\begin{document}

\title{RADIANCE: Radio-Frequency Adversarial Deep-learning Inference for Automated Network Coverage Estimation}

\author{\IEEEauthorblockN{Sopan Sarkar, Mohammad Hossein Manshaei, and Marwan Krunz}
\IEEEauthorblockA{Department of Electrical and Computers Engineering, University of Arizona, Tucson, USA \\
\{sopansarkar, manshaei, krunz\}@arizona.edu}
}

\maketitle

\begin{abstract}
Radio-frequency coverage maps (RF maps) are extensively utilized in wireless networks for capacity planning, placement of access points and base stations, localization, and coverage estimation.
Conducting site surveys to obtain RF maps 
is labor-intensive and sometimes not feasible.
In this paper, we propose \textit{radio-frequency adversarial deep-learning inference for automated network coverage estimation} (RADIANCE), a generative adversarial network (GAN) based approach for synthesizing RF maps 
in indoor scenarios.
RADIANCE utilizes a semantic map, a high-level representation of the indoor environment to encode 
spatial relationships and attributes of objects within the environment and guide the RF map generation process. 
We introduce a new gradient-based loss function that computes the magnitude and direction of change in received signal strength (RSS) values from a point within the environment.
RADIANCE incorporates this loss function along with the
antenna  pattern to capture signal propagation 
within a given indoor configuration and generate new patterns
under new configuration, antenna (beam) pattern, 
and center frequency.
Extensive simulations are conducted to compare
RADIANCE with ray-tracing simulations of RF maps.
Our results show that RADIANCE achieves a mean average error (MAE) of 0.09, root-mean-squared error (RMSE) of 0.29, peak signal-to-noise ratio (PSNR) of 10.78, and multi-scale structural similarity index (MS-SSIM) of 0.80. 
\end{abstract}


\section{Introduction}
\label{sec_Introduction}
Radio-frequency coverage maps (RF maps) are vital in wireless communication, finding applications in network and capacity planning, interference coordination, resource allocation, localization, handoff management, and coverage estimation \cite{7849214}. 
However, collecting RF data and conducting site surveys is complicated \cite{6216368, 8695724} due to the complex and dynamic nature of the wireless environment and the wide range of factors that can affect signal propagation, such as building materials, terrain, and interference.
For example, as stated in \cite{niu2018recnet}, for omnidirectional communications, it took the authors about 11.9 hours to coarsely survey an area of $4500~\mbox{meter}^2$ using only 475 reference points.

With the introduction of millimeter-wave (mmW) and sub-Terahertz (sub-THz) frequencies in 5G and beyond 5G (B5G) communication systems, RF map generation has become even more challenging.
The signal propagation at these frequencies suffers from high atmospheric attenuation, limited penetration, susceptibility to shadowing, and inability to operate in non-line-of-sight (NLOS) scenarios~\cite{rangan2014millimeter}.
While atmospheric attenuation can be compensated for by using high-dimensional antenna arrays and beamforming techniques, blockage remains a key problem, making cell coverage highly dependent on the environment.
Consequently, with the slightest changes in the environment and communication parameters, there are new scenarios to be investigated, which cannot be easily solved by the existing site survey techniques \cite{6216368}.

One potential solution to address the above challenges is to use ray tracing-based computer simulation to generate RF maps. 
Ray tracing techniques can accurately solve electromagnetic equations in complex environments \cite{ray_tracing_complexity}, albeit at the expense of computational resources \cite{romero2022radio}. 
Moreover, the efficacy of ray tracing relies on precise representations of the propagation environment, including 3D models of all objects and obstacles and their corresponding electromagnetic properties \cite{romero2022radio}. 
Therefore, there is a need for a simple, fast, and cost-effective solution to generate RF maps.

To address these challenges, researchers have proposed various methods such as crowdsourcing \cite{wu2017,yin2017} and machine learning \cite{9247298}.
Crowdsourcing involves collecting data from various user devices, which can be used to generate RF maps.
However, aggregating data from heterogeneous devices is difficult, which may result in erroneous decisions.
On the other hand, traditional machine learning techniques such as generalized linear models and k-nearest neighbors cannot provide fast and accurate results and may require the precise engineering of very complex networks. 

Generative models have gained significant attention in recent times as a method of synthesizing datasets.
Several extensions of generative adversarial network (GAN) models were applied to solve wireless communication problems. 
For instance, GAN-based models have been proposed for channel modeling~\cite{yang2019generative}, beamforming~\cite{liu2023full},  anomaly detection~\cite{zhou2021radio}, signal spoofing~\cite{shi2019generative},
and wideband channel estimation~\cite{balevi2021wideband}.
In particular, GANs have been employed to generate additional measurements with improved diversity, thus expanding the training dataset, reducing data collection time, and saving human effort~\cite{zou2020}. 

In~\cite{Kim2020}, Kim et al. proposed a conditional GAN (cGAN) that extracts part of an RF map as a window, learns the indoor space in partitions, and combines the RF maps of various access points to generate a complete map.
In \cite{zou2020}, Zou et al. use a mobile robot to construct an RF map and estimate the RSS for new coordinates using Gaussian process regression conditioned least-squares-based GAN.
Liu et al. \cite{Liu2020} determined the best Radio dot location using a dimension-aware conditional GAN with modified loss functions and a multi-stage training strategy from radio heatmap data collected by human experts.
In~\cite{Njima2021}, Njima et al. proposed a GAN for RSS data augmentation for indoor localization, which requires a small set of experimentally collected labeled data. 
The authors in \cite{supreme} proposed Supreme, a fine-grained RF map reconstruction scheme based on crowd-sourced data.
They model a residual block (ResBlk) to explore spatial-temporal relationships in historical coarse-grained RF maps and build a real-time fine-grained RF map using deep spatial-temporal reconstruction networks.
\begin{figure}[t!]
 \begin{subfigure}[b]{0.24\textwidth}
    \centering
    \label{fig:gan}
    \includegraphics[scale=0.23]{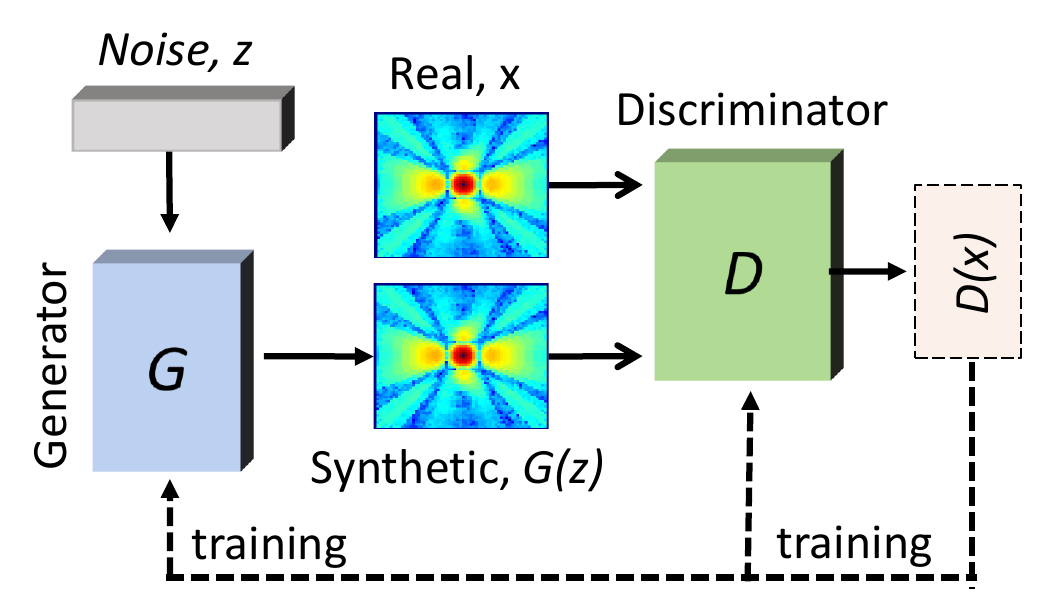}
    \caption{Basic GAN}
    \end{subfigure}%
  \begin{subfigure}[b]{0.24\textwidth}
    \centering
    \label{fig:cgan}
    \includegraphics[scale=0.23]{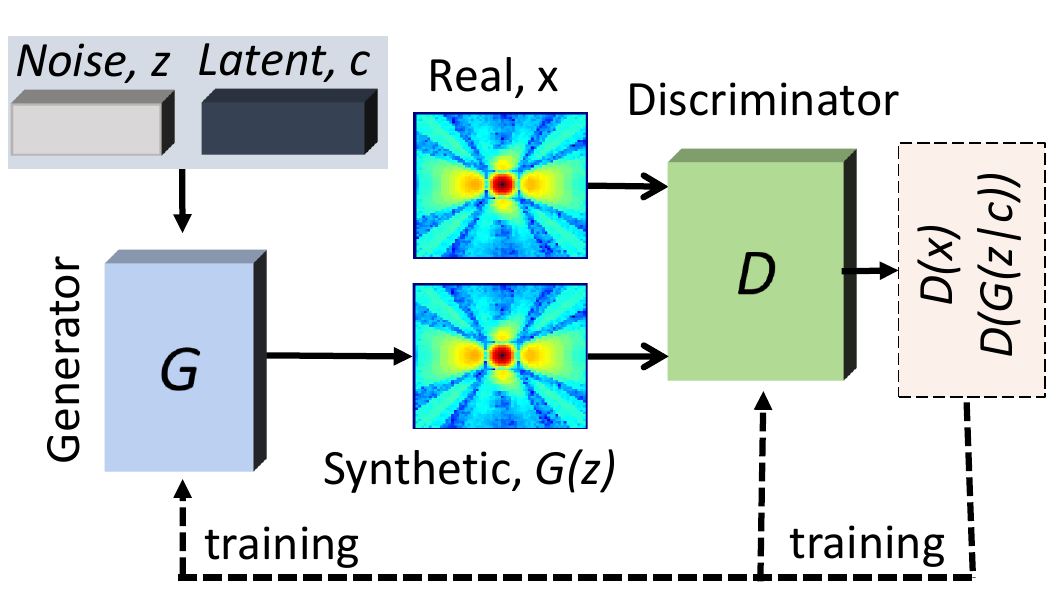}
    \caption{Conditional GAN}
    \end{subfigure}%
\caption{Two variants of GAN.}
\label{fig:GAN_Models}
\end{figure}

In all the above works, a complete RF map was constructed based on incomplete data for a specific environment, BS location, center frequency, and antenna configuration.
They fail to generate RF maps for a completely new environment or different network configurations.
Thus it limits the ability of the algorithms to adapt to changes in the environment as new sets of data need to be collected every time.
To address this limitation, we propose a novel approach called RADIANCE (Radio-frequency Adversarial Deep-learning Inference for Automated Network Coverage Estimation), which uses a modified cGAN structure to generate synthetic RF maps for any new indoor scenario and network configuration without the need for collecting new data.
Specifically, RADIANCE uses a semantic map of the indoor radio environment that provides both spatial relationships and attributes of the objects within the environment, such as room geometry, position and type of materials, and their shapes.
We also propose a gradient-based loss function that captures the signal propagation from the BS by computing the magnitude and direction of the changes of RSS at each point in the given indoor environment.
RADIANCE incorporates this loss function along with the semantic map of the indoor environment and the antenna pattern of the transmitting BS, enabling easy and cost-effective generation of realistic RF maps. 
We compare the synthetic RF maps generated by RADIANCE with RF maps generated using ray tracing-based simulation.
We observe that RADIANCE generates highly representative RF maps for any new floor plan, antenna configurations, and BS location while considering the specific center frequency.
Moreover, compared to the ray tracing simulated RF maps, RADIANCE achieves a mean average error (MAE) of 0.09, root-mean-squared error (RMSE) of 0.29, peak signal-to-noise ratio (PSNR) of 10.78 and multi-scale structural similarity index (MS-SSIM) of 0.80.

The subsequent sections of this paper are structured as follows. Section~\ref{sec_Preliminary} provides an overview of various GAN networks. 
Section~\ref{sec_RADIANCE} introduces the RADIANCE model and loss functions.
Section~\ref{sec_Eval} presents the description of the dataset and the experimental evaluation of RADIANCE followed by conclusions in Section~\ref{sec_Conclusion}.
\begin{figure}[t!]
 \begin{subfigure}[b]{0.24\textwidth}
    \centering
    \frame{\includegraphics[scale=0.5]{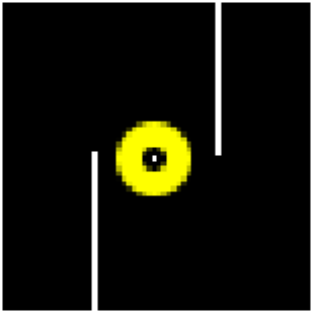}}
    \caption{Semantic map}
    \label{fig:semantic_map}
    \end{subfigure}%
  \begin{subfigure}[b]{0.24\textwidth}
    \centering
    \includegraphics[scale=0.44]{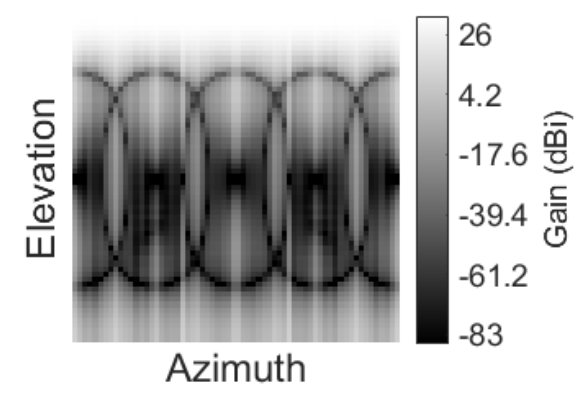}
    \caption{Antenna pattern}
    \label{fig:antenna_pattern}
    \end{subfigure}%
\caption{(a) Semantic map with white representing brick walls, black representing concrete floor, and a yellow circle representing the BS location, and (b) $4\times4$ UPA antenna pattern in azimuth and elevation.}
\label{fig:maps}
\end{figure}
\begin{figure*} 
 \begin{subfigure}[]{0.25\textwidth}
    \centering
    \label{fig:Arch}
    \includegraphics[scale=0.31]{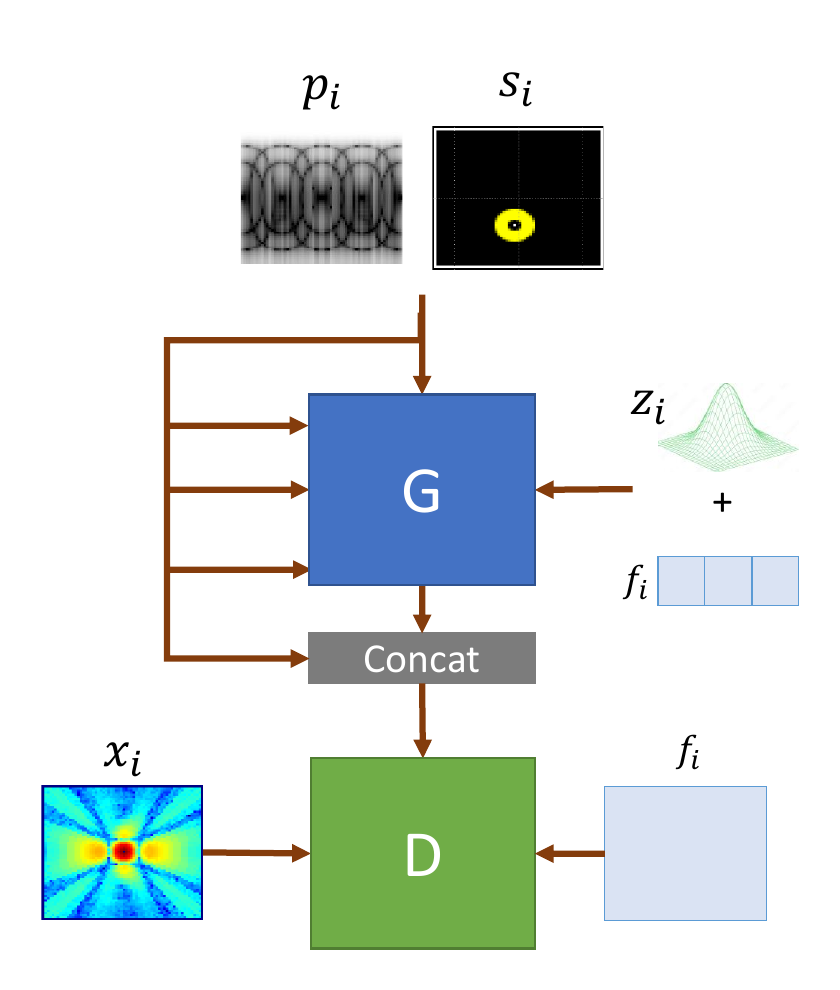}
    \caption{RADIANCE architecture}
    \end{subfigure}%
  \begin{subfigure}[]{0.25\textwidth}
    \centering
    \label{fig:ResBlk}
    \includegraphics[scale=0.31]{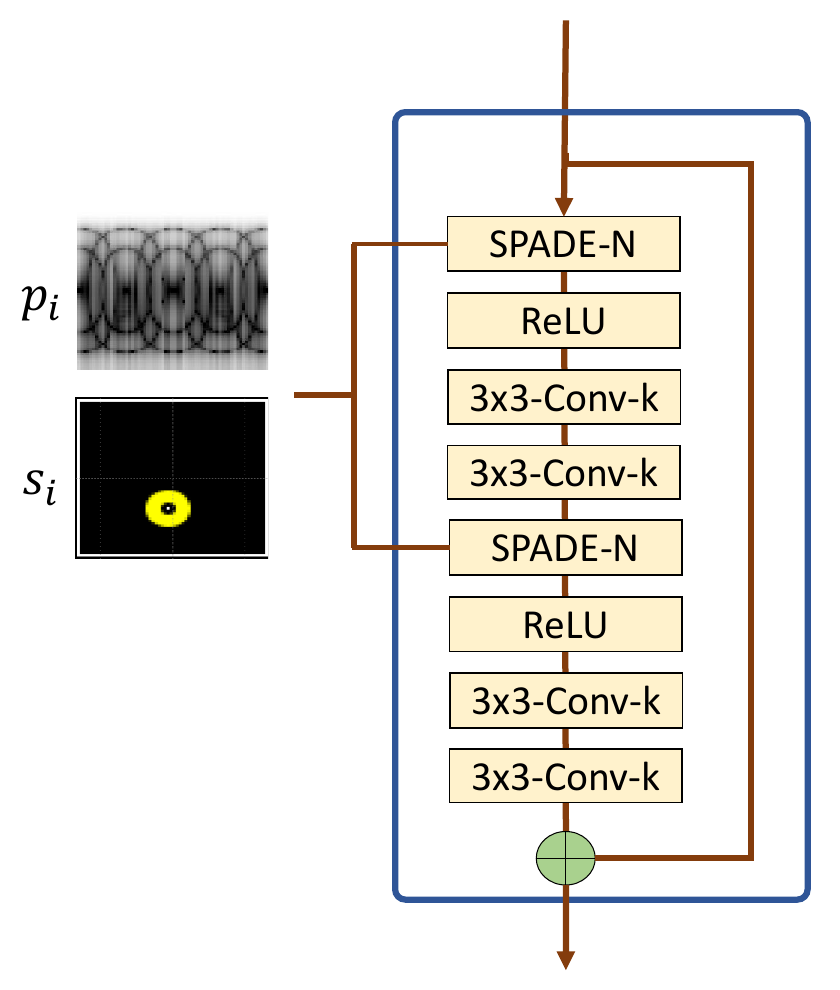}
    \caption{Residual block}
    \end{subfigure}%
    \begin{subfigure}[]{0.25\textwidth}
      \centering
    \label{fig:Generator}
    \includegraphics[scale=0.31]{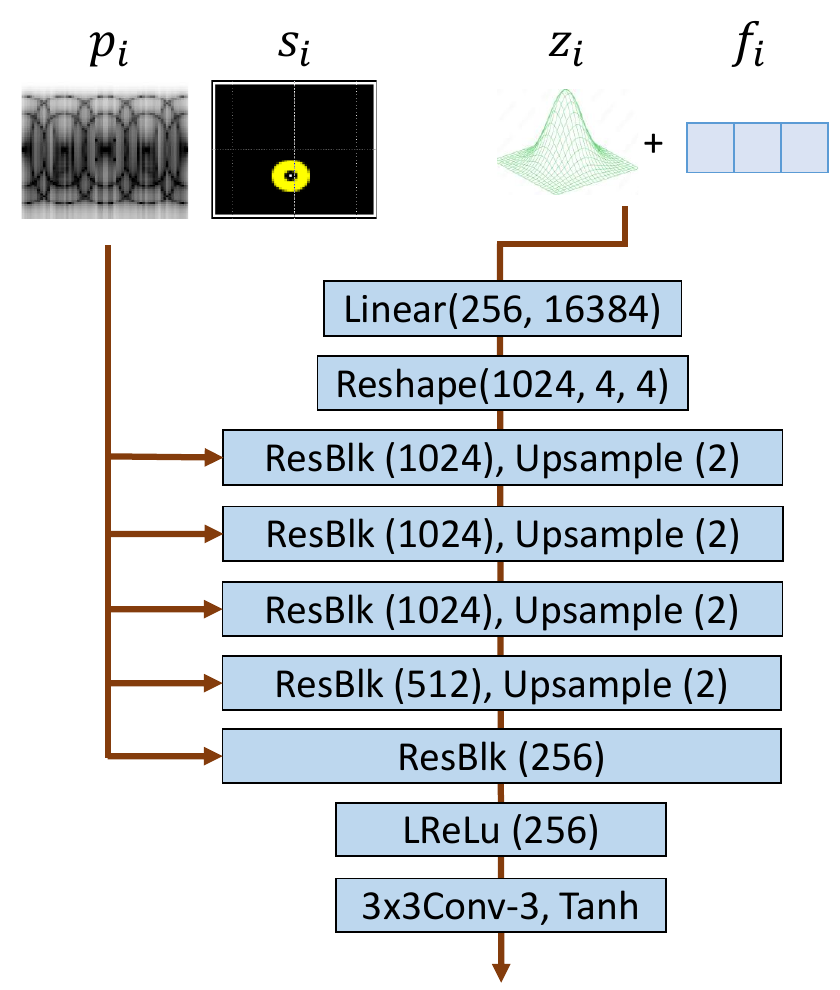}
    \caption{Generator network}
    \end{subfigure}%
    \begin{subfigure}[]{0.25\textwidth}
      \centering
    \label{fig:Disc}
    \includegraphics[scale=0.31]{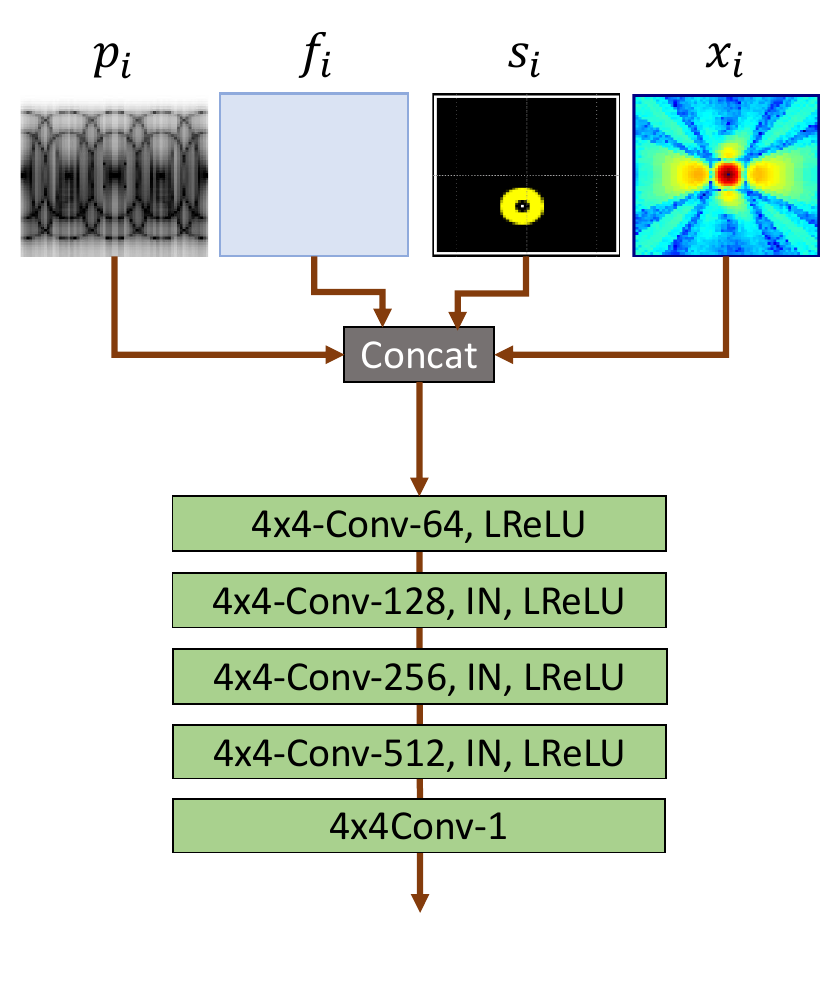}
    \caption{Discriminator network}
    \end{subfigure}%
\caption{(a) Architecture of RADIANCE, including generator and discriminator network connections, (b) design of 
the residual block used in constructing the generator, (c) generator, and  (d) discriminator architecture in RADIANCE.}
\label{fig:RADIANCEModel}
\end{figure*}
\section{Generative Adversarial Networks}
\label{sec_Preliminary}
A basic GAN consists of two neural networks~\cite{goodfellow2014generative}: A generator ($G$) and a discriminator ($D$), as depicted in Figure~\ref{fig:GAN_Models}(a).
$G$ and $D$ are trained simultaneously via an adversarial learning technique, which is essentially a two-player zero-sum min-max game played between $G$ and $D$.
Once the GAN has been trained, the game converges to a Nash equilibrium, where $G$ and $D$ cannot independently increase their payoff.
At this convergence point, the min-max and maxi-min values are equal to $-\log4$~\cite{goodfellow2014generative}. 
The objective function of a basic GAN can be formalized using the Jensen-Shannon divergence metric as follows:
\begin{multline}
	\min\limits_{G} \max\limits_{D} V(G, D)
	=\min\limits_{G} \max\limits_{D} \Big(  \mathbb{E}_{x \sim p_{data}(x)} [\log(D(x))]\\
	+\mathbb{E}_{z \sim p_{z}(z)} [\log(1-D(G(z)))\Big).
	\label{eq:minimaxlogGAN}
\end{multline}
Here, $x$ is real data samples and $z$ is noise.
$G(z)$ is a function that transforms a noise distribution ($p_{z}$) into an estimated data distribution ($p_{g}$).
$D(x)$ maps the input data distribution $p_{data}$ to the range of $[0,1]$, indicating the likelihood of a sample being real and not generated by the generator.

In our design, we employ a specific type of GAN called cGAN to synthesize RF maps (see Figure~\ref{fig:GAN_Models}(b)). 
cGANs are powerful models that enable the conditional generation of data. 
In addition to $z$, cGAN takes in extra conditioning information $c$, such as class labels or auxiliary input, to guide the generation of samples~\cite{mirza2014conditional}.
In other words, a cGAN learns disentangled representations $p_{data}$ by utilizing $c$ as a conditioning factor for $G$. 
The objective function of a cGAN model is as follows:
\begin{multline}
	\min\limits_{G} \max\limits_{D} V(G, D)
	=\min\limits_{G} \max\limits_{D} \Big( \mathbb{E}_{x \sim p_{data}(x)} [\log(D(x|c))]\\
	+\mathbb{E}_{z \sim p_{z}(z)} [\log(1-D(G(z|c)))]\Big).
	\label{eq:cGAN}
\end{multline}
cGAN is highly customizable, as one can learn the expected properties by combining various loss functions into the objective function and using corresponding conditions as input~\cite{Pathak_2016_CVPR}.

\section{RADIANCE Design}
\label{sec_RADIANCE}
In this section, we present the design of the proposed RADIANCE.
We first formulate the problem, then discuss the basic structure of RADIANCE in detail, and finally explain the various loss functions used to modify the GAN objective.
\subsection{Problem Setup}
\label{sec_problem_formulation}
We focus on synthesizing new RF maps for indoor scenarios.
Each floor plan $i$ is characterized by an RF map $x_i \in \mathbb{R}^{l \times w\times3}$ and a semantic map $s_i \in \mathbb{R}^{l \times w\times3}$, where $l$ and $w$ is the length and width of the RF maps.
$s_i$ provides information about the dimension, type, and location of objects within the floor plan $i$. 
As shown in Fig.~\ref{fig:maps}(a), the semantic map represents a square-shaped room and reveals three features: a white structure indicating brick walls, a black background indicating a concrete floor, and a yellow circle indicating the BS location.
Each RF map is characterized by the antenna pattern $p_i \in \mathbb{R}^{l \times w\times1}$ used at the BS.
The antenna pattern is represented by the antenna gain values for various azimuth and elevation angles.
An example of the antenna pattern for a $4\times4$ UPA is shown in Fig.~\ref{fig:maps}(b).
In order to incorporate the influence of transmission frequency, we employ a one-hot encoded categorical vector $f_i$ of length $k$. This vector assigns a value of $1$ to the specific center frequency, while all other values remain $0$.
Thus the RF map $\mathbf{x}_i$ of each floor plan is conditioned by
\begin{equation}
    c_i = \{s_i, p_i, f_i\}.
\end{equation}
Our goal is to train a generative model $G(.)$ to synthesize RF maps for new and never before seen floor plans, BS location, and antenna configuration given center frequency, i.e.,
\begin{equation}
    \Tilde{x}_i = G(z | c_i) \in \mathbb{R}^{l \times w \times 3}.
\end{equation}
\subsection{RADIANCE Architecture}
\label{sub_rad_arc}
Fig.~\ref{fig:RADIANCEModel}(a) presents the architecture of RADIANCE, which is composed of two types of neural networks: (i) a generator $G$ and (ii) a discriminator  $D$. 
The generator of RADIANCE is depicted in Fig.~\ref{fig:RADIANCEModel}(c), along with the type of layers and their associated hyperparameters.
The main building block of the generator is a modified residual block (ResBLK) which consists of a spatially-adaptive normalization layer \cite{Park_2019_CVPR} (SPADE-N) followed by a rectified linear (ReLU) layer and stacks of two convolutional (Conv) layers.
We use upsampling to upsample the input after each ResBlK.
Finally, the generator produces RF maps given the input $z$ and $c_i$.

We use a PatchGAN \cite{Isola_2017_CVPR} discriminator with instance normalization (IN), and leaky ReLu (LReLu) activation function as shown in Fig.~\ref{fig:RADIANCEModel}(d).
The discriminator tries to differentiate between real and synthetic RF maps.
The input to the discriminator is a concatenation of the semantic map, the antenna pattern, the real or synthetic RF map, and a one-hot encoded matrix.
This one-hot encoded matrix has a dimension of $l\times w \times k$ and is generated from the one-hot encoded vector $f_i$.
The output of the discriminator is a $5\times5$ patch matrix indicating whether the image patch is real or fake.



\begin{figure*}[ht]
    \centering
 \begin{subfigure}[b]{0.175\textwidth}
   \centering
    \label{fig:sq1_HM}
    \includegraphics[scale=0.45]{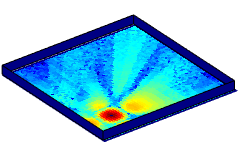}
    \caption{Room 1}
    \end{subfigure}%
  \begin{subfigure}[b]{0.175\textwidth}
    \centering
    \label{fig:sq2_HM}
    \includegraphics[scale=0.45]{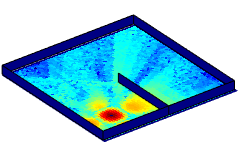}
    \caption{Room 2}
    \end{subfigure}%
  \begin{subfigure}[b]{0.175\textwidth}
    \centering
    \label{fig:sq3_HM}
    \includegraphics[scale=0.45]{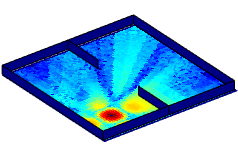}
    \caption{Room 3}
    \end{subfigure}%
  \begin{subfigure}[b]{0.175\textwidth}
    \centering
    \label{fig:sq4_HM}
    \includegraphics[scale=0.45]{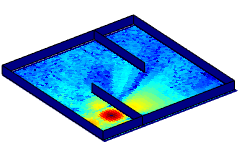}
    \caption{Room 4}
    \end{subfigure}%
  \begin{subfigure}[b]{0.175\textwidth} 
    \centering
    \label{fig:lshaped_HM}
    \includegraphics[scale=0.45]{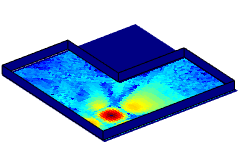}
    \caption{L-shaped room}
    \end{subfigure}
    \begin{subfigure}[b]{0.1\textwidth} 
    \centering
    \label{fig:colorbar}
    \includegraphics[scale=0.25]{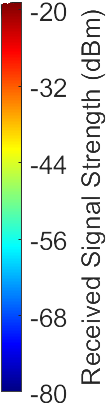}
    \end{subfigure}
\caption{3D view of the different floor plans along their corresponding RF map for $4\times4$ UPA at 28 GHz.}
\label{fig:floor_plans}
\end{figure*}

\subsection{Loss Function}
\label{sub_loss_function}


\textbf{\textit{Mean absolute error loss}}: Mean absolute error loss ($L_{\textit{\mbox{MAE}}}$) captures the average difference between corresponding pixel values and provides a straightforward and interpretable measure of dissimilarity between $x_i$ and $\Tilde{x}_i$.
For example, for $n$ generated samples, it is represented as:
\begin{equation}
    L_{\textit{\mbox{MAE}}}= \frac{1}{n} \sum_{1}^n | x_i - \Tilde{x}_i|
\end{equation}

\textbf{\textit{Focal loss}}: 
In RF maps, pixels representing high RSS values are crucial in applications like localization and radio network planning.
Focal loss $L_{\textit{\mbox{FL}}}$ compels the model to assign greater importance to important pixels in the RF map, thereby amplifying the influence of high RSS value pixels on the training process and the final decision.

\textbf{\textit{Feature matching loss}}: To ensure intermediate features align properly, the feature matching loss ($L_{\textit{\mbox{FM}}}$) incentivizes $G$ to grasp the high-level structure and semantics of the ground truth data. 
This promotes the generation of visually captivating and realistic samples. 
$L_{\textit{\mbox{FM}}}$ can be computed as follows:
\begin{equation}
    L_{\textit{\mbox{FM}}} = \frac{1}{m} \sum_{i=1}^{m} \{||Q^{x_i} - Q^{\Tilde{x_i}}||^2_2\}.
\end{equation}
Here, $m$ denotes the number of feature maps, and $Q^{x_i}$ and $Q^{\Tilde{x_i}}$ represents features extracted from $x_i$ and $\Tilde{x}_i$.
We use $D$ to extract the feature maps.

\textbf{\textit{VGG Feature matching loss}}: VGG feature matching loss ($L_{\textit{\mbox{VGG}}}$) is a variant of $L_{\textit{\mbox{FM}}}$ that leverages features extracted by a VGG network \cite{Simonyan15} for GAN training. 
By using $L_{\textit{\mbox{VGG}}}$, $G$ is incentivized to produce samples that not only deceive $D$ at a high-level feature representation level but also capture the intricate details and textures found in real samples. 

\textbf{\textit{Gradient loss}}: By inspecting the RF maps, we observe that the distribution of the RSS depends on the antenna pattern.
The signal propagates smoothly along the antenna's boresight, but there are abrupt changes in the RSS between the boresight and the null direction.
To capture these effects, we use a gradient loss ($L_{\textit{\mbox{GL}}}$).
Basically, we use Sobel filer to obtain the gradient of $x_i$ and $\Tilde{x}_i$, i.e., $\nabla x_i$ and $\nabla \Tilde{x}_i$.
The magnitude of the gradient tells us how quickly the RSS is changing, while the direction of the gradient tells us the direction in which the RSS is changing most rapidly.
To measure $L_{\textit{\mbox{GL}}}$, we determine the changes in the magnitude and direction between $\nabla x_i$ and $\nabla \Tilde{x}_i$ using KL-divergence (KL) and cosine similarity (CS), respectively.
Thus we can write $L_{\textit{\mbox{GL}}}$ as
\begin{equation}
    L_{\textit{\mbox{GL}}} = \frac{1}{n} \sum_{1}^{n} \Big( \mbox{KL}(\nabla  x_i, \nabla  \Tilde{x}_i) + \mbox{CS}(\nabla  x_i, \nabla  \Tilde{x}_i) \Big)
\end{equation}

Based on the above loss functions, we can rewrite the total generator loss function as follows:
\begin{multline}
    L_\textit{\mbox{Total}} = L_\textit{\mbox{G}} + \lambda_\textit{\mbox{mae}} L_\textit{\mbox{MAE}} + \lambda_\textit{\mbox{fl}} L_\textit{\mbox{FL}}\\ + \lambda_\textit{\mbox{fm}} L_\textit{\mbox{FM}} + \lambda_\textit{\mbox{vgg}} L_\textit{\mbox{VGG}} + \lambda_\textit{\mbox{gl}} L_\textit{\mbox{GL}},
\end{multline}
where $\lambda_\textit{\mbox{mae}}, \lambda_\textit{\mbox{fl}},\lambda_\textit{\mbox{fm}}, \lambda_\textit{\mbox{vgg}},~\mbox{and}~\lambda_\textit{\mbox{gl}}$ are the weights of the corresponding loss functions.

\begin{table}[hb]
    \caption{Simulation Parameters}
    \centering
    \begin{tabular}{|c|c|c|}
    \hline
      {\bf Parameter}  & {\bf Description} & {\bf Value} \\
    \hline
    \hline
Frequency   & sub-6 GHz                 & 5 GHz\\
                & mmWave                & 28, 70 GHz \\
    \hline
    Wall        & Height                & 4 meters \\
                & Material              & Brick \\
    \hline
    Floor       & Material              & Concrete \\
    \hline
    Base station & Transmit Power        & 0 dBm\\
                & Height                & 3 meters \\
    \hline
    BS antenna  & Uniform planer array  & $4\times4$\\
                &                       & $6\times6$\\
                &                       & $8\times8$\\
                &                       & $10\times10$\\
                &                       & $12\times12$\\
    \hline
    Receiver    & Height                & 1.5 meters\\
                & Antenna               & Isotropic antenna\\
    \hline
    Floor Plans & Dimensions            & $10 \times 10$ meters \\
    \hline
    \end{tabular}
    \label{tab:simu_para}
\end{table}
\begin{figure*}[!hbt]
    \centering
    \includegraphics [scale=0.54]{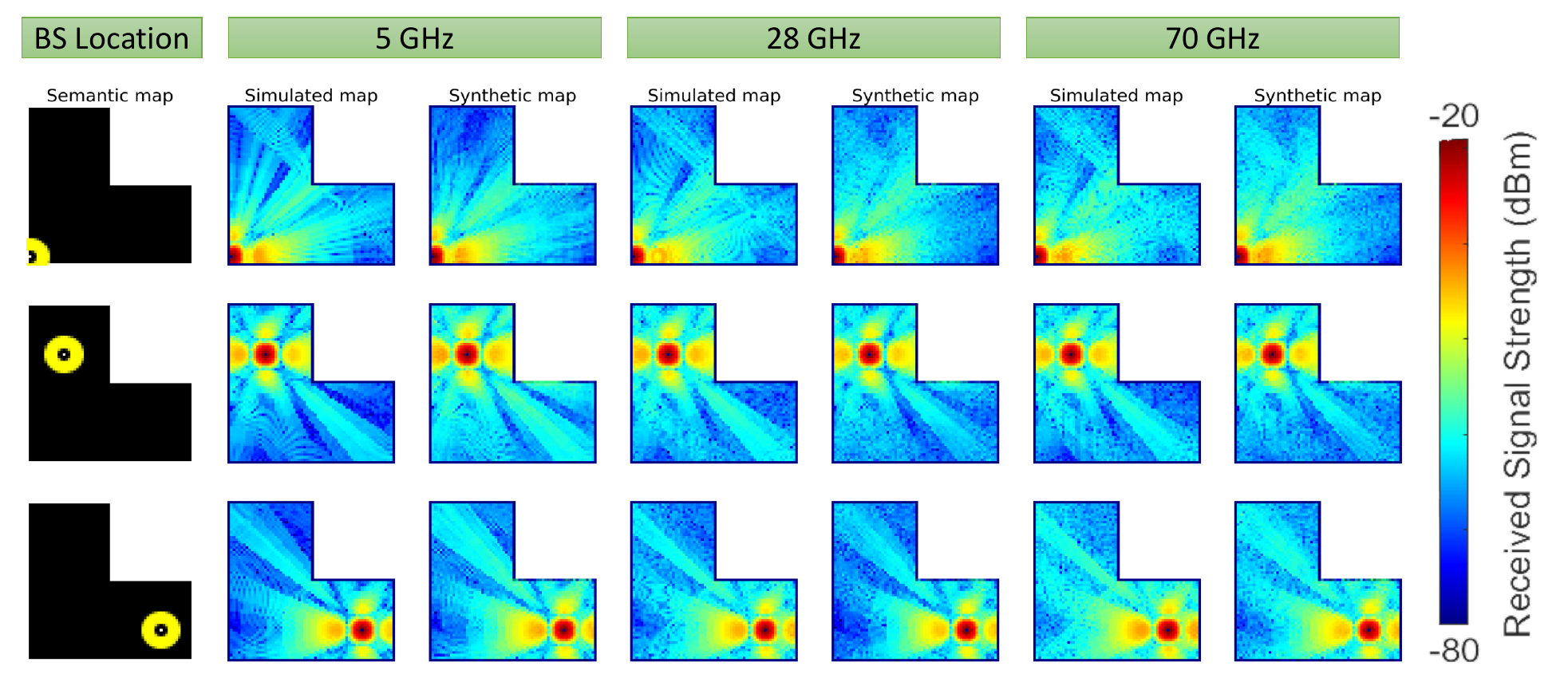}
    \caption{Comparison between ray tracing simulated and RADIANCE synthesized RF maps for L-shaped room at different BS locations and center frequencies. The BS antenna is a  $4\times4$ UPA.}
    \label{fig:results1}
\end{figure*}
\begin{figure*}[!hbt]
    \centering
    \includegraphics [scale=0.56]{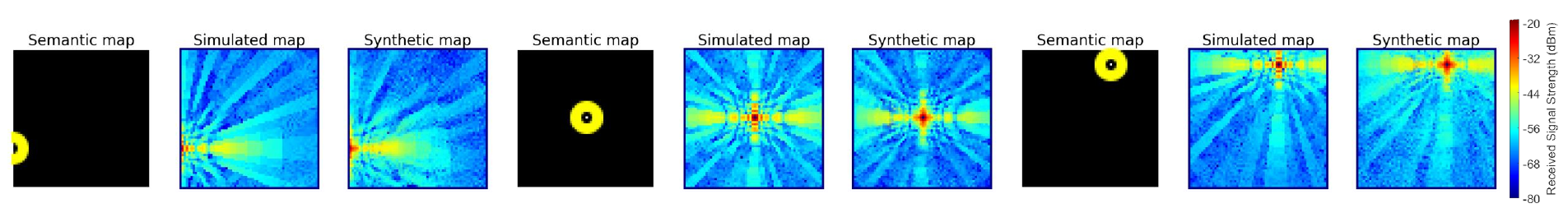}
    \caption{Comparison between simulated and synthetic RF maps generated using a $10\times10$ UPA at 28 GHz for different BS locations in Room 1.}
    \label{fig:results2}
\end{figure*}
\section{Experimental Results}
\label{sec_Eval}
In this section, we first describe our dataset and then provide a detailed report on the performance of RADIANCE in synthesizing new RF maps. 
\subsection{Dataset Generation}
\label{sec_Dataset}
It is imperative to note that obtaining practical measurements for generating ground truth RF maps is crucial for validating the accuracy of RADIANCE.
In this study, we rely on ray tracing simulations as they are highly accurate \cite{ray_tracing_complexity} and can be used to obtain multiple coverage maps. 

In this paper, we employ MATLAB-based ray tracing simulations to generate RF maps for training and validating the RADIANCE model.
Specifically, we use the shooting and bouncing ray (SBR) method~\cite{ling1989shooting} to generate the propagation paths between the BS and multiple receivers (RX). 
We generate RF maps for various center frequencies, antenna patterns, and BS locations, considering five different indoor scenarios. 
These scenarios have a consistent dimension of $10\times10$ meters but differ in their geometries.
The floor plans, illustrating the layouts of these scenarios along with their corresponding RF maps for a $4\times4$ UPA at 28 GHz, are depicted in Figure~\ref{fig:floor_plans}(a-e). 
We assume the walls are constructed of brick material, and the floor is made of concrete. 
These parameters can be easily adjusted and modified as necessary.


In order to generate an RF map, we divide the room area into a $64\times64$ grid, with each grid of size 0.024 $\mbox{meter}^2$ for all the floor plans.
We consider that each RX is equipped with an isotropic antenna with an omnidirectional pattern.  
In contrast, we employ uniform planar arrays (UPA) with patch antenna elements at the BS.
The dimensions of the UPAs considered are $4\times4$, $6\times6$, $8\times8$, $10\times10$, and  $12\times12$.
Moreover, we consider three carrier frequencies: 5 GHz, 28 GHz, and 70 GHz.
A single BS is placed 3 meters above the ground in each grid tile, with the antenna's broadside facing downwards along the Z-axis, from the ceiling towards the floor. 
We place a grid of RXs occupying all grid tiles in the floor plan.
The RX grid is placed at a height of 1.5 meters. 
Generally, indoor BSs do not employ beam tracking; thus, we maintain a fixed BS beam direction. 
In other words, we let the antenna beam's boresight align with the antenna array's broadside.
The RSS values are measured for the RX grid and are used to generate a coverage heatmap.
Table~\ref{tab:simu_para} provides the values of the key configuration parameters, including operating frequencies, material types, BS/RX configurations, etc.
The dataset consists of 74,752 samples in total.

\subsection{Evaluation}
\label{sec_Evaluation}
GANs can generate realistic samples by accurately modeling complex multi-dimensional data. 
However, assessing the performance of GANs is difficult, and various evaluation techniques have been proposed in the literature.
Most of these techniques rely on visual inspection of the synthesized data samples, while others, such as the RMSE and MS-SSIM, quantitatively compare GAN-generated data with the ground truth. 
In this section, we evaluate the performance of RADIANCE in synthesizing new RF maps.

We evaluate the performance of RADIANCE in performing two different tasks.
For Task 1, we fix the antenna pattern to $4\times4$ UPA and want RADIANCE to generate RF maps for completely new floor plans and BS locations, for different center frequencies.
In Task 2, we fix the floor plan and the center frequency and want RADIANCE to synthesize RF maps for completely new antenna patterns and BS locations.
For each task, we first visually compare the RADIANCE-generated synthetic RF maps with the ray trace simulated maps and then compute the MAE, RMSE, PSNR, and MS-SSIM to compare them quantitatively.

For Task 1, we select Rooms 1 to 4 (see Fig.~\ref{fig:floor_plans}) and their corresponding features as the training set.
For the test set, we consider the L-shaped room and its corresponding features.
As seen in Fig.~\ref{fig:results1}, for different BS locations (depicted by the yellow circle in the semantic map) and center frequencies, RADIANCE was able to generate representative RF maps for an L-shaped floor plan which has a completely different geometry than Rooms 1 to 4.
Moreover, for Task 1, RADIANCE achieves MAE of 0.06, RMSE of 0.23, PSNR of 12.81, and MS-SSIM of 0.91.
The results are provided in Table.~\ref{tab:sim_resutls}.
The best value one can expect from this comparison is also represented in column 2 in the table. 

For Task 2, we fix the room geometry to Room 1 and the center frequency to 28 GHz for both the training and test set.
But for the training set, we consider different UPAs of dimension $4\times4$, $6\times6$, $8\times8$, and $12\times12$ UPA.
As for the test set, we only consider $10\times10$ UPA.
The goal is to see if RADIANCE can generate RF maps for completely new antenna configurations at different BS locations within the floor plan.
Fig.~\ref{fig:results2} depicts the results for Task 2.
The synthetic RF maps are visually similar to the simulated RF maps.
In terms of MAE, RMSE, RSNR, and MS-SSIM, it achieves a value of 0.13, 0.36, 8.75, and 0.70, respectively.
\begin{table}[hbt!]
    \caption{RADIANCE Performance}
    \centering
    \begin{tabular}{|c|c|c|c|c|}
    \hline
    {\bf Parameter}  & {\bf Best Value}  & {\bf Task 1}  & {\bf Task 2}   & {\bf Avg.} \\
    \hline
    \hline
    MAE     &0  &0.06   &0.13   &0.09\\
    \hline
    RMSE    &0   &0.23  &0.36   &0.29\\
    \hline
    PSNR    &$\inf$  &12.81    &8.75   &10.78\\
    \hline
    MS-SSIM &1   &0.91  &0.70   &0.80\\
    \hline
    \end{tabular}
    \label{tab:sim_resutls}
\end{table}

\section{Conclusion}
\label{sec_Conclusion}
This paper proposed RADIANCE, a conditional GAN-based method for synthesizing RF maps for given wireless communication attributes. 
The framework incorporated modified neural network architectures for the generator and discriminator.
Moreover, a new gradient-based loss function was proposed to capture signal propagation effects. 
The synthesized RF maps resemble those produced by ray tracing simulations, as confirmed by both visual and quantitative evaluations.

RADIANCE currently supports one BS with specific floor
plan dimensions and center frequencies, limiting its generalizability. Thus, restrictions exist concerning dimension, frequency flexibility, and number of BS. Our future work includes
verifying the model with different experimental datasets,
modifying the architecture to accommodate different floor
plan dimensions and center frequencies, and investigate the
impact of multiple BS and furniture orientation within rooms.
\bibliographystyle{IEEEtran}
\bibliography{RADIANCE}
\end{document}